\documentclass{article}
\usepackage{spconf,amsmath,graphicx}
\usepackage{graphicx}
\usepackage{amsmath}
\usepackage{subfigure}
\usepackage{multirow}
\usepackage{amssymb}
\usepackage[strings]{underscore}

\title{Learning Granularity-Aware Convolutional	Neural Network for Fine-Grained Visual
Classification}
\name{Jianwei Song, Ruoyu Yang}
\address{National Key Laboratory for Novel Software Technology\\
Nanjing University\\
Nanjing, China\\}
\begin{document}
%
\maketitle
\begin{abstract}
Locating discriminative parts plays a key role in
fine-grained visual classification due to the high similarities between different objects. Recent works based on convolutional neural networks utilize the feature maps taken from the last convolutional layer to mine discriminative regions.
However, the last convolutional layer tends to focus on the whole object due to the large receptive field, which leads to a reduced ability to spot the differences.
To address
this issue, we propose a novel Granularity-Aware
Convolutional Neural Network (GA-CNN) that progressively explores discriminative features. Specifically, GA-CNN utilizes the differences of the receptive fields at different layers to learn multi-granularity features, and it exploits larger granularity information based on the smaller granularity information found at the previous stages.
To further boost the performance, we introduce an object-attentive module that can effectively localize the object given a raw image.
GA-CNN does not need bounding boxes/part annotations and can
be trained end-to-end. Extensive experimental results show that our approach achieves state-of-the-art performances on three benchmark datasets.
\end{abstract}

\section{Introduction}
Fine-grained visual classification (FGVC) focuses on
distinguishing subtle visual differences within a basic-level
category (e.g., species of birds\cite{CUB}). In recent years, generic object recognition has achieved great success due to convolutional neural networks (CNNs)\cite{ResNet}. However, FGVC is still a challenging task where discriminative parts are too subtle to be well captured by traditional CNNs. Learning discriminative feature representations from distinguishable parts plays a key role in FGVC. Existing methods can
be roughly divided into two categories.
The first category often consists of two different subnetworks. Specifically, a localization subnetwork is designed to localize discriminative parts and a classification subnetwork is followed to learn powerful feature representations from these parts. Earlier works\cite{Part-RCNN,Deep-LAC} belonging to this category utilize the bounding boxes/part annotations to capture visual details in local regions. However, collecting extra annotated information is labor-intensive and requires professional knowledge, making these methods less practical. Hence, weakly supervised FGVC methods \cite{MA-CNN,RA-CNN,MAMC,MGE} have been proposed, these methods use attention mechanisms instead of extra annotations to localize discriminative regions.
The second category often enhances mid-level feature learning by encoding higher-order information. The most classic approach is B-CNN\cite{BCNN}, which performs outer product on two feature maps taken from different branches. Because the result of B-CNN is high dimensional, low-rank bilinear pooling\cite{LR-BCNN} is proposed to reduce feature dimensions before conducting the bilinear operation.

However, all the aforementioned methods utilize the feature maps taken from the last convolutional layer to mine discriminative regions.
We argue that the last convolutional layer tends to focus on the whole object due to the large receptive field,
but there are high similarities between different objects, which leads to a reduced ability to spot the discriminative parts. Meanwhile, the receptive fields of neurons at the earlier layers are relatively small, so these neurons can capture part regions intrinsically. Based on this assumption,
we propose a novel Granularity-Aware Convolutional Neural Network (GA-CNN), which utilizes the differences of the receptive fields at different layers to learn multiple granularity-specific features progressively.
To further boost the performance, we introduce an object-attentive module that utilizes an attention map to estimate the bounding box of the object region given a raw image. GA-CNN does not need bounding boxes/part annotations, and state-of-the-art performances are reported on
three benchmark datasets.
\begin{figure*}
\centering
\includegraphics[width=0.9\textwidth]{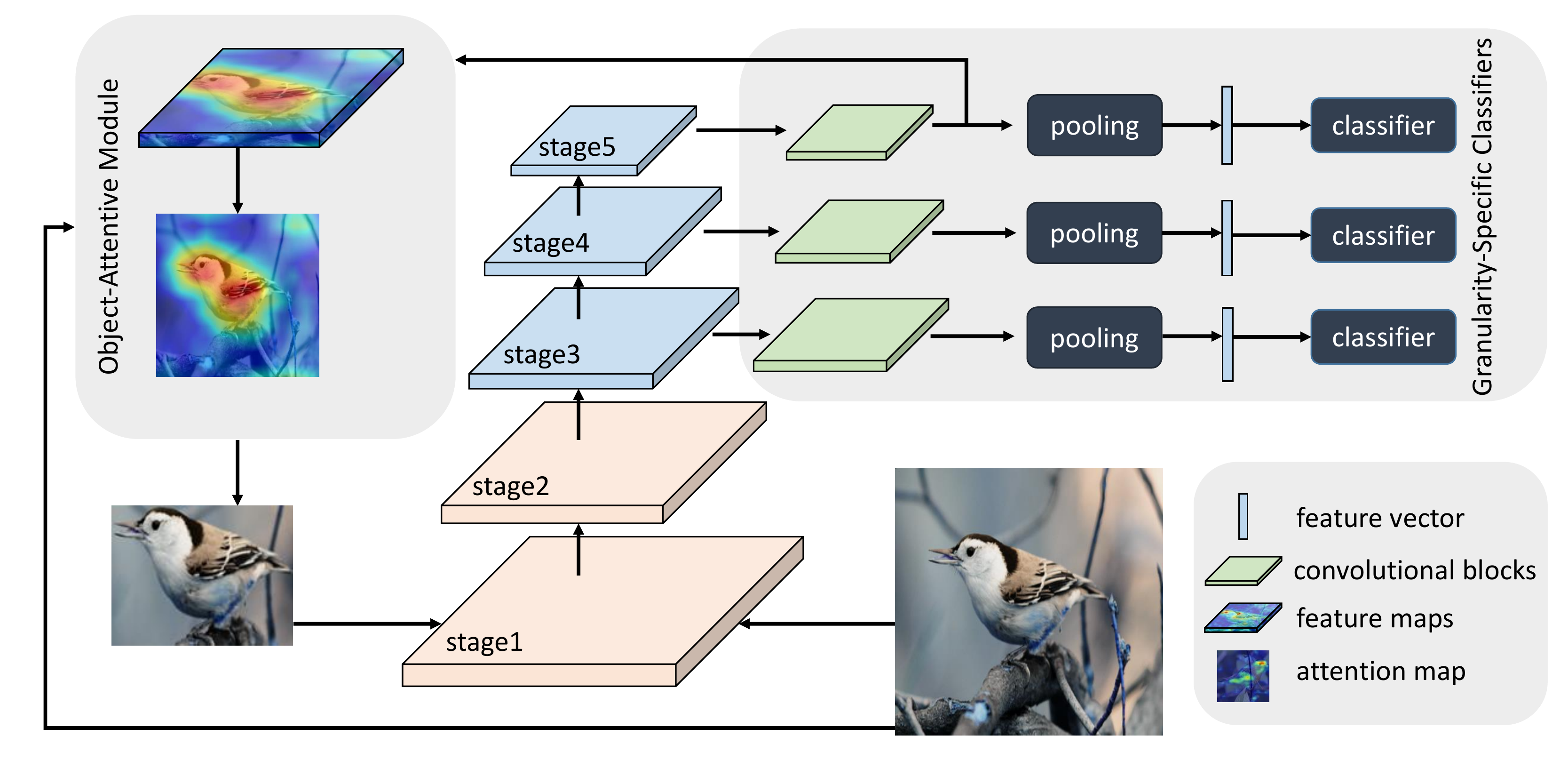}
\caption{The framework of granularity-aware convolutional neural network (GA-CNN).}
\label{fig:framework}
\end{figure*}
\section{Related Works}
\subsection{Visual Attention for FGVC}
MA-CNN\cite{MA-CNN} proposed to learn attentions by
grouping semantically similar channels. RA-CNN\cite{RA-CNN} proposed to locate the object and discriminative parts recursively. MAMC\cite{MAMC} applied the multi-attention multi-class constraint in a metric learning framework to mine part attentions. CIN\cite{CIN} proposed to model channel interaction to mine semantically
complementary information.
Our method is to learn feature representations at different granularities, which is significantly different from previous methods.
\subsection{Feature Extraction}
SSD\cite{SSD} and FPN\cite{FPN} have made great success in the object detection field by extracting features from different layers.
\cite{CrossX} proposed a cross-layer regularizer to improve the robustness of feature representations by matching the prediction distributions across multiple layers. \cite{HCNN} integrated hierarchical convolutional activations from different layers via kernel fuse to model part interaction.
Our approach also involves extracting features from different layers but explores multi-granularity discriminative parts within an object
instead of localizing multi-scale objects within an image as in \cite{SSD,FPN}.

\section{Our Approach}
In this section, we detail the proposed granularity-aware convolutional neural network (GA-CNN). An overview of GA-CNN is displayed in Figure \ref{fig:framework}.
We use Resnet50\cite{ResNet} for illustration, but our method is not limited to any specific network.
\subsection{Learning Granularity-Specific Features}
Let $B$ be our backbone feature extractor, which has $L$ stages.
We denote the feature maps extracted from $B$ at the $i^{th}$ stage as $B_{(i)}\in R^{c_i\times w_i\times h_i}$, where $w_i, h_i, c_i$ represents the width, height, and the number of channels of the feature maps respectively.
The feature granularity of $B_{(i)}$ increases as the receptive field increases.
We enhance the $B_{(i)}$ with $1\times 1$ and $3\times 3$ convolutional blocks and pool it into a feature vector $V_{(i)}$ with GTKP\cite{ELoPE}.
We introduce a classifier $cls_i$ for feature vector $V_{(i)}$ and the classification loss is the cross-entropy loss:
\begin{equation}
L_{cls}^i = -log(p_i(l)), \quad p_i = cls_i(V_{(i)})
\end{equation}
where $l$ is the ground-truth label of the input image, $p_i$ is the prediction probability.
Our objective is to optimize the feature representations at the last $S$ stages, and the final loss is:
\begin{equation}
L = \sum_{i\in [L-S+1,\cdots,L]}L_{cls}^i
\end{equation}
During training,
the lower stages will be forced to mine discriminative local regions due to the limited semantic information and receptive field.
Meanwhile,
the parameters of the $i^{th}$ stage are not only updated by classifier $cls_i$ for better classification, but also updated by classifiers $\{cls_{i+1},\cdots,cls_{L}\}$ for providing better intermediate feature representations.
With this simple strategy, the network will exploit larger granularity information based on the smaller granularity information found at the previous stages, and
we can make use of the earlier layers which have relatively small receptive fields to spot small regions and the deep layers which have rich semantic information to guarantee the discrimination of these small regions.
When learning multiple granularity-specific classifiers (GSC) at different stages,
our model can progressively obtain multi-granularity feature representations.

\begin{figure}
\centering
\includegraphics[width=\linewidth]{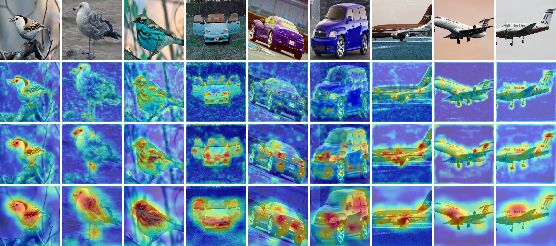}
\caption{Visualization of feature maps. The first row is the raw input images. The attention maps from the second to the fourth rows correspond to the third to the fifth stages of Resnet50. [Best viewed in color]}
\label{fig:vis}
\end{figure}

\textbf{Justification}:	We visualize the feature maps taken from the last three stages of Resnet50\cite{ResNet} equipped with GSC to justify our idea. As shown in Figure \ref{fig:vis}, taking the bird as an example, the attention map in the second row focuses on the relatively small granularity features such as the beak, the third row focuses on the relatively large granularity features such as the head, and the attention map in the last row focuses on the whole body. The visualization proves that our model focuses on discriminative parts at different granularities.
\subsection{Object-Attentive Module}
Inspired by MGE-CNN\cite{MGE}, we introduce an object-attentive module (OAM) to further boost the classification accuracy, which makes feature learning and object localization in a mutually reinforced way. Supposing that the feature maps for object localization are $F$, we first create an attention map $A$ by taking
the only channel with the largest activation of $F$.
We then normalize the attention map $A$ by scaling the value between 0 and 1:
\begin{equation}
A^* = \dfrac{A-\text{min}(A)}{\text{max}(A) - \text{min}(A)}
\end{equation}
After that, we can get a clipping mask $C$ by setting $A^*{(i,j)}$ which is greater than threshold $\alpha$ to 1, and others to 0:
\begin{equation}
C_{(i,j)} =
\begin{cases}
1,\quad\text{if }A^*{(i,j)} > \alpha \\
0,\quad\text{otherwise}.
\end{cases}
\end{equation}
where $\alpha$ is a hyper-parameter.
By taking the smallest rectangle that can cover the whole positive regions of $C$, we can obtain a coarse bounding box. We then crop the corresponding region from the raw image and feed it into the network to get fine-grained feature representations.
\section{Experiments}
\begin{table}
\caption{Three fine-grained benchmark datasets.}
\label{table:dataset}
\centering
\begin{tabular}{|c|c|c|c|c|}
\hline
Dataset & Name & \#Class & \#Train & \#Test \\ \hline
CUB-200-2011 & Bird & 200 & 5,994 & 5,794 \\ \hline
FGVC-Aircraft & Aircraft & 100 & 6,667 & 3,333 \\ \hline
Stanford Cars & Car & 196 & 8,144 & 8,041 \\ \hline
\end{tabular}
\end{table}
\begin{table}[t]
\caption{Comparison results with the state-of-the-art methods. ``-" means the result is not reported in the relevant paper.}
\label{table:sota}
\centering
\begin{tabular}{ccccc}
\hline
Methods &Backbone
&Bird &Aircraft
& Car \\ \hline
RA-CNN\cite{RA-CNN} & VGG & 85.3 & 88.1 & 92.5 \\
MA-CNN\cite{MA-CNN} & VGG & 86.5 & 89.9 & 92.8 \\
\hline
MAMC\cite{MAMC} & Resnet50 & 86.2 &- &92.8 \\
NTS\cite{NTS} & Resnet50 & 87.5 & 91.4 & 93.3 \\
API-Net\cite{API-NET} & Resnet50 & 87.7 & 93.0 & \textbf{94.8} \\
Cross-X\cite{CrossX} & Resnet50 & 87.7 & 92.6 & 94.5 \\
DCL\cite{DCL} & Resnet50 & 87.8 & 93.0 & 94.5 \\
CIN\cite{CIN} & Resnet50 & 87.5 & 92.6 & 94.1 \\
ISQRT-COV\cite{iSQRT-COV} & Resnet50 & 88.1 & 90.0 & 92.8 \\
MGE-CNN\cite{MGE} & Resnet50 & 88.5 &- & 93.9 \\
S3N\cite{S3N} & Resnet50 & 88.5 & 92.8 & 94.7 \\
Ours &Resnet50 & \textbf{90.4} &\textbf{93.3} & \textbf{94.8} \\ \hline

MAMC\cite{MAMC} & Resnet101 & 86.5 &- &93.0 \\
CIN\cite{CIN} & Resnet101 & 88.1 &92.8 &94.5 \\
API-Net\cite{API-NET} & Resnet101 & 88.6 &93.4 &\textbf{94.9} \\
ISQRT-COV & Resnet101 & 88.7 &91.4 &93.3 \\

MGE-CNN\cite{MGE} & Resnet101 & 89.4 & - &93.6 \\
Ours &Resnet101 & \textbf{90.1}&\textbf{94.1} &\textbf{94.9} \\ \hline

API-Net\cite{API-NET} & Densenet161 & 90.0 & 93.9 & \textbf{95.3} \\
Ours &Densenet161 & \textbf{90.5}&\textbf{94.1} &95.1 \\ \hline
\end{tabular}
\end{table}
\subsection{Implementation Details and Datasets}
Our approach is flexible and can be implemented on various convolutional neural networks. We validate the performances of our model on Resnet50\cite{ResNet}, Resnet101\cite{ResNet}, and Densenet161\cite{DenseNet} which are all pre-trained on the ImageNet dataset. The input image size is $448\times 448$ as most state-of-the-art methods. We extract feature maps at different granularities from the last three stages and utilize the enhanced feature maps taken from the last stage to locate the object.
At the inference phase, given a raw image, we first obtain coarse predictions and an object bounding box, then crop the corresponding region from the raw image and feed it into the network to obtain fine-grained predictions, and the final classification result is the average of the coarse and fine-grained predictions of all classifiers. The hyper-parameter $\alpha$ is set to 0.3.
Our model is optimized by Stochastic Gradient Descent with the momentum of 0.9, epoch number of 200, weight decay of 0.00001, mini-batch of 20. The learning rate of the backbone layers is set to 0.002, and the newly added layers are set to 0.02. The learning rate is adjusted by the cosine anneal scheduler\cite{Warm}. We use PyTorch to implement our experiments.

To show the efficiency of our method, we conduct experiments on three fine-grained datasets: CUB-200-2011\cite{CUB}, FGVC-Aircraft\cite{CRAFT} and Stanford Cars\cite{CAR}.
The detailed information of each dataset is summarized
in Table \ref{table:dataset}.

\subsection{Comparisons with State-of-the-Art Methods}
We compare our approach with state-of-the-art weakly-supervised methods\cite{iSQRT-COV,MAMC,MA-CNN,RA-CNN,MGE,NTS,S3N,API-NET,CrossX,DCL,CIN} on three benchmark datasets. The comparison results are summarized in Table \ref{table:sota}. Our approach achieves state-of-the-art on CUB-200-2011, with the Resnet50 backbone, our model can surpass all other methods even equipped with more advanced backbones by large margins. Our approach gets the best result on FGVC-Aircraft compared with other methods under the same backbone. For the Stanford Cars dataset, our methods implemented on the Resnet50 and Resnet101 achieve the best result like API-Net\cite{API-NET}. However, API-Net\cite{API-NET} spots discriminative regions by comparing image pairs,
it needs to consider different pairwise image combinations within a mini-batch
and requires large computing resources.
\begin{table}
\caption{Contribution of each proposed component.}
\label{table:each}
\centering
\begin{tabular}{cccc}
\hline
Methods & GSC & OAM & ACC(\%) \\ \hline
Resnet50 & & & 83.7 \\
Resnet50 & \checkmark & & 88.7 \\
Resnet50 & \checkmark & \checkmark & 90.4 \\ \hline
\end{tabular}
\end{table}
\begin{table}
\caption{Ablation studies on the threshold $\alpha$.}
\label{table:alpha}
\centering
\begin{tabular}{c|ccccc}
\hline
$\alpha$ & 0.1 & 0.2 & 0.3 & 0.4 & 0.5\\ \hline
Accuracy & 89.64 & 89.85 & 90.38 & 89.92 & 89.96\\
\hline
\end{tabular}
\end{table}
\begin{table}[t]
\caption{Ablation studies on the pooling methods.}
\label{table:k}
\centering
\begin{tabular}{c|ccc}
\hline
Methods &GAP &GMP & GTKP\\ \hline
Accuracy &88.23 &90.09 & 90.38\\\hline
\end{tabular}
\end{table}
\begin{figure}[t]
\centering
\includegraphics[width=\linewidth]{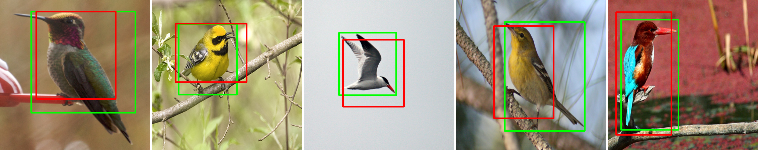}
\caption{Visualization of the ground-truth and the estimated bounding boxes of the object region. [Best viewed in color]}
\label{fig:loc}
\end{figure}
\subsection{Analysis and Discussion}
To well understand our method, we conduct comprehensive ablation studies on
CUB-200-2011 dataset with Resnet50 as the backbone.

\textbf{The effect of granularity-specific classifiers}:
As shown in Table \ref{table:each}, when introducing multiple granularity-specific classifiers (GSC) into Resnet50, the model obtains 88.7\% accuracy which is 5.0\% higher than the Resnet50 baseline. This result indicates the effectiveness of this component.

\textbf{The effect of object-attentive module}: As shown in Table \ref{table:each}, when introducing the object-attentive module (OAM),
our model obtains 90.4\% accuracy which is 1.7\% higher than Resnet50 with only granularity-specific classifiers.
We also visualize the bounding box estimated by our model in CUB-200-2011 dataset. As shown in Figure \ref{fig:loc}, we use green and red rectangles to denote the ground-truth and the estimated bounding boxes respectively, the object-attentive module can effectively localize the object region.

\textbf{The effect of the threshold $\alpha$}:
The classification results concerning the threshold $\alpha$ are shown in Table \ref{table:alpha}. We can see that our model is not sensitive to $\alpha$, it can adaptively adjust the attention map to fit different values of $\alpha$.

\textbf{The effect of pooling methods}:
We show ablation studies on pooling methods: GAP (global average pooling), GMP (global max pooling), and GTKP (global top-k pooling)\cite{ELoPE}. The classification results are shown in Table \ref{table:k}. GTKP obtains the best result which is 2.15\% and 0.29\% higher than GAP and GMP respectively. GTKP takes the top-k activations into account to capture multiple discriminative parts but GMP only utilizes the maximum activation, which indicates that learning multiple discriminative parts is favorable to FGVC.
GAP obtains the worst result because it averages all the activations and introduces lots of confusing features.
\section{Conclusion}
In this paper, we propose a novel granularity-aware convolutional neural network (GA-CNN) for fine-grained visual classification, which learns robust multi-granularity feature representations and makes feature learning and object localization in a mutually reinforced way. The proposed network does not need bounding boxes/part annotations and can be trained in an end-to-end way. Experiments are comprehensively conducted on three benchmark datasets and state-of-the-art performances are achieved. In the future, we will investigate how to model the interaction among multi-granularity features to capture more subtle visual differences.

\end{document}